\documentclass[10pt,twocolumn,letterpaper]{article}

\usepackage{cvpr}
\usepackage{times}
\usepackage{epsfig}
\usepackage{graphicx}
\usepackage{amsmath}
\usepackage{amssymb}
\usepackage{mathtools}
\usepackage{enumitem}
\usepackage{flushend}
\usepackage{tgcursor}
\usepackage{multirow}
\usepackage{makecell}
\usepackage{booktabs}
\usepackage[hyphens]{url}

\usepackage[pagebackref=true,breaklinks=true,letterpaper=true,colorlinks,bookmarks=false]{hyperref}

\newcolumntype{L}[1]{>{\raggedright\let\newline\\\arraybackslash\hspace{0pt}}m{#1}}
\newcolumntype{C}[1]{>{\centering\let\newline\\\arraybackslash\hspace{0pt}}m{#1}}
\newcolumntype{R}[1]{>{\raggedleft\let\newline\\\arraybackslash\hspace{0pt}}m{#1}}

\cvprfinalcopy 

\def\cvprPaperID{8455} 

\ifcvprfinal\fi
\begin{document}

\title{Learning Transformation-Aware Embeddings for Image Forensics}

\author{Aparna Bharati$^1$, Daniel Moreira$^1$, Patrick Flynn$^1$, Anderson Rocha$^2$, Kevin Bowyer$^1$, Walter Scheirer$^1$\\
\(^1\) University of Notre Dame, IN, USA \hspace{0.5cm}
\(^2\) University of Campinas, SP, Brazil\\
}

\maketitle

\begin{abstract}
   A dramatic rise in the flow of manipulated image content on the Internet has led to 
   an aggressive response from the media forensics research community. New efforts have incorporated increased usage of techniques from computer vision and machine learning to detect and profile the space of image manipulations. 
   This paper addresses Image Provenance Analysis, which aims at discovering relationships among different manipulated image versions that share content. One of the main sub-problems for provenance analysis that has not yet been addressed directly is the edit ordering of images that share full content or are near-duplicates. The existing large networks that generate image descriptors for tasks such as object recognition may not encode the subtle differences between these image covariates. This paper  
   introduces a novel deep learning-based approach to provide a plausible ordering to images that have been generated from a single image through transformations. Our approach learns transformation-aware descriptors using weak supervision via composited transformations and a rank-based quadruplet loss. To establish the efficacy of the proposed approach, comparisons with state-of-the-art handcrafted and deep learning-based descriptors, and image matching approaches are made. Further experimentation validates the proposed approach in the context of image provenance analysis.

\end{abstract}

\section{Introduction}
\label{sec:intro}

\begin{figure}[t]
\vspace{-0.3cm}
\begin{center}
\includegraphics[width=0.99\linewidth]{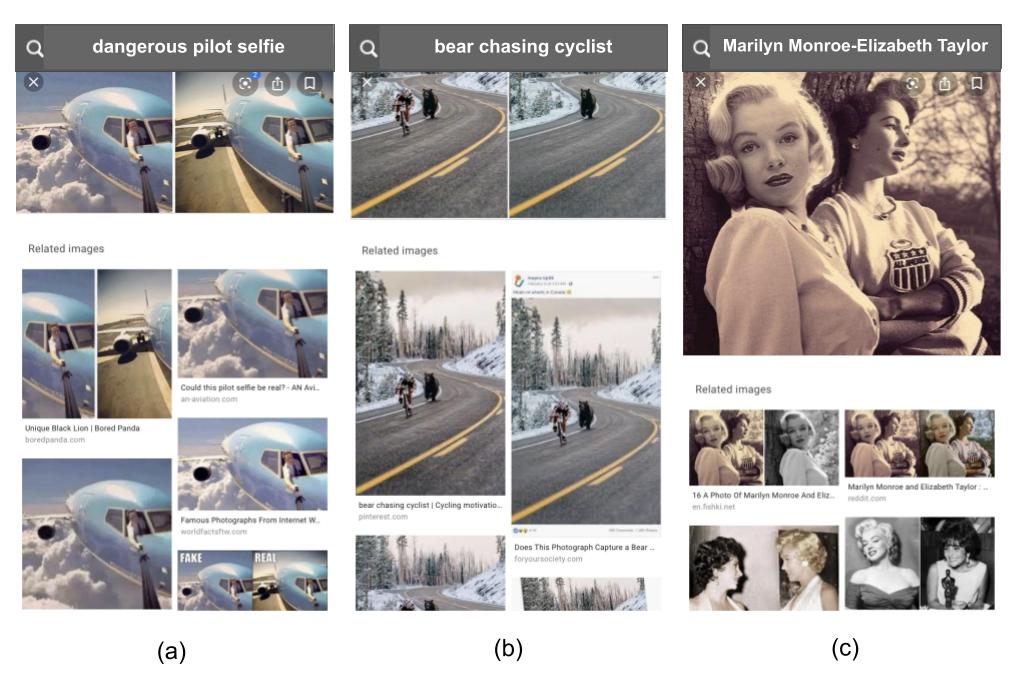}
\end{center}
\vspace{-0.5cm}
   \caption{An image search on the web using keywords related to some fake photos that went viral (Google search keywords: most+viral+fake+photo) reveals many versions of the same photo in the related images section. Most of these images are transformed versions of the original. For example, some cropped and perspective transformed versions appear in the related images for query (a), a sheared image version appears for (b) and some color transformed images appear in the related section of query (c). Given multiple variants for any given image, an important question is: can we find the original? Not necessarily just for maliciously manipulated images, this question is valid for a number of circumstances where image manipulation is present. We propose a framework that helps determine the ordering of image transformations to trace the provenance of the content.}

\label{fig:teaserreal}
\vspace{-5mm}
\end{figure}

In the fight against the spread of disinformation~\cite{cdc, conv} through manipulated media content~\cite{wsj}, understanding the story and intent behind the manipulated object is critical. Whether the manipulations are benign or malicious, a step-by-step analysis of how the current version of the manipulated image or video was generated helps us in answering more holistic and contextual questions than just whether a given image or video is real or fake. Unlike the early days of the web, a media object today does not exist in isolation. Most often there are multiple versions of a single image online at any given time (Fig.~\ref{fig:teaserreal}). Tracing the uploads, downloads and re-uploads of original image content with modifications can help assess the reach of that content. If an image is a composite, analyzing its different versions 
and other images that donated content to its creation can provide explanations for the types of manipulations and their complexity. Beyond images on social media, we can also consider other image-based applications such as autonomous driving, where  it is important to know if the data being used to train a visual recognition system is the original version and distinguish between benign and malicious data attacks. 

\begin{figure}[t]
\begin{center}
\includegraphics[width=1.0\linewidth]{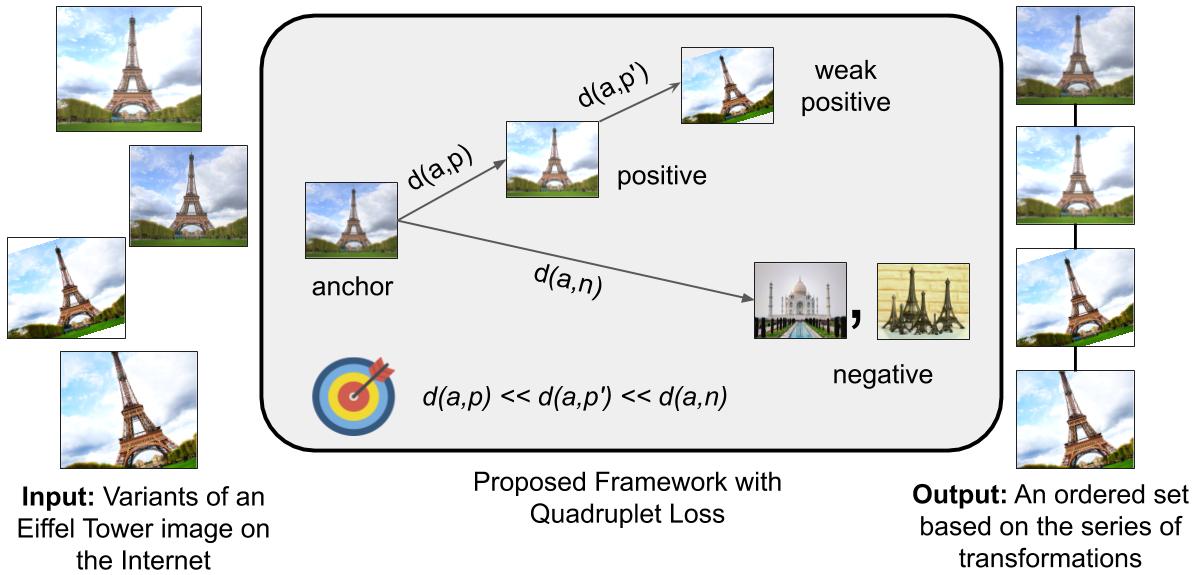}
\end{center}
\vspace{-0.2cm}
   \caption{An example of image ordering in the context of image provenance analysis. Given a set of near-duplicate images, an ordered set where the order explains the process of transforming one image to the other is desired as output. The figure includes a schematic representation of the proposed technique used to achieve this. The goal is to order relative distances between compositely-transformed images through the use of a ranking-based quadruplet loss function.}

\label{fig:teaser}
\vspace{-5mm}
\end{figure}

\textit{Image Provenance Analysis}~\cite{moreira2018image} aims to understand the evolution of a media object in question by establishing pairwise associations among images related in content to generate its provenance graph. An example of provenance analysis for a small set of images is shown in Figure~\ref{fig:teaser}. Provenance analysis usually involves two stages: (1) a specialized image retrieval stage to obtain images related to a query called \textit{provenance filtering}, and (2) a \textit{graph construction} stage to model the relationships between the retrieved images. While both steps employ vision-based techniques (they rely on establishing image correspondences), there has been limited research on the latter problem. 

In the literature, provenance graph construction algorithms are based on image phylogeny algorithms~\cite{dias2012image, dias2013toward, oliveira2014multiple}. Provenance graph construction uses generalized definitions of image relationships and circumvents the phylogeny constraints such as specific image formats and a two-donor limit for composite images~\cite{bharati2017uphy}. These algorithms use points of interest to establish image correspondence and compute pairwise image dissimilarity by describing matched local regions. Upon computing a dissimilarity score for all possible pairs, a greedy algorithm is employed to create a minimally connected Directed Acyclic Graph (DAG)~\cite{moreira2018image} of images. Depending on the type of manipulations performed on an image, the graph can be very complicated with multiple donor images (images that share partial content) and long chains formed by near-duplicate images (images derived from a single image through a series of transformations). This work focuses on improving the fidelity of reconstructing the chains in the provenance graphs. 

Greedy graph algorithms treat dissimilarity values as adjacency weights and rely heavily on invariant image matching. Image matching is a common task in many computer vision problems such as object recognition, 3D reconstruction, scene understanding and image retrieval. 
A desired property for representations used to solve these problems is invariance to view changes, compression and other image transformations. Optimizing this property while learning the mapping between the data $\mathbb{X}$ and label $\mathbb{Y}$ can easily miss understanding the fine differences between image transformations.
Thus invariance becomes something of a misnomer in the context of forensics.

We propose to learn representations that are aware of different versions of images in the transformation space, and depending on the number of transformations, can encode appropriate distance among near-duplicate images. 
This approach can be useful for improving dissimilarity computation between different versions of an image.
Better understanding of the subtle differences in the variants can lead to improved rank-based output in deducing a sequence of transformations for image forensics~\cite{farid2009image} and cultural analytics~\cite{manovich2009cultural}. It can also be used to define acceptable standards of edited data for learning algorithms. To our knowledge, this is the first work that focuses on solving the image ordering aspect of provenance analysis. This work also highlights the importance of awareness of transformation-based differences among near-duplicate images while learning image representations.

\vspace{-0.2cm}
\section{Related Work}
\label{sec:rw}

Most of the recent research related to image provenance analysis and other media forensics problems~\cite{moreira2018image, huh2018fighting, zhou2018learning, d2018patchmatch, wu2018busternet, rossler2018faceforensics} is related to the DARPA Media Forensics program~\cite{darpa} through its challenges and datasets.
The first end-to-end algorithm~\cite{moreira2018image} proposed for provenance analysis coming out of that program followed a two-step strategy to obtain a graph-based relationship representation of images that are near-duplicates of or share partial content with the query image. Firstly, a specialized image retrieval algorithm that employed Speeded Up Robust Features (SURF)~\cite{bay2006surf} for description, Optimized Product Quantization (OPQ)~\cite{ge2013optimized, johnson2017billion} for efficient indexing, and iterative filtering is used to obtain a list of images related to the query. 
In the second step, a 
keypoint-based pairwise image comparison is performed to obtain a dissimilarity matrix. The most feasible provenance graph is created using this matrix through a hierarchical clustering-based graph expansion method.
The latter step can also be interpreted as ordering pair similarities between multiple image pairs, and is therefore a natural extension to pairwise image comparison. 
For output, the ordered pairings are modeled as a graph, where each edge denotes a transformation-based correspondence, and the images in the pair are the 
vertices~\cite{bharati2017uphy}. 

Image ordering can be defined as a specific task that outputs an ordered list of provided input images along with producing a matching score between two images. This can correspond to ranking images with respect to distance from a query image in retrieval algorithms or identification scenarios in recognition. Approaches in the existing literature which are a source of inspiration for the proposed work are discussed below. We categorize them broadly to explain how advances in techniques for these general tasks become useful in designing a pipeline to learn transformation-aware embeddings for ordering images in provenance analysis.

\subsection{Image Matching}
Ordering can be considered as an additional step to image matching as it involves a measure of how well images match. Image matching algorithms use points of interest to learn whether two images have the same content with respect to structure~\cite{bay2006surf, yi2016lift, leng2018local}. 
Such algorithms are integral parts of correspondence tasks where one cares about the exact locations two image contents match.
Due to the nature of the applications that require image correspondence (dense or sparse), these matches~\cite{zagoruyko_2015, revaud2016deepmatching} are intended to encode invariance across a range of transformations, \textit{i.e.}, they attempt to compute a feature space where different versions of an image region map to the same vicinity. 
In the case of handcrafted feature matching, the design of keypoint detectors and patch descriptors such as SIFT~\cite{lowe1999object} and SURF~\cite{bay2006surf} 
mostly imparts invariance to linear affine transformations such as scale, transformation, rotation and others. 

For learning-based correspondence, invariance to more transformations can be achieved through hierarchical convolutional architectures and by introducing transformed or augmented versions of data points during training~\cite{xuzhang2017learningdisc, fischer2014descriptor}. Encoding invariance in one or the other way helps to improve the robustness of matching and make it stable under extreme transformations or view-point changes. For applications such as provenance analysis, we require an understanding of the measure of invariance  (\textit{i.e.}, how much invariance is present) in addition to mapping features of near-duplicate or similar images closer to each other than those of dissimilar images.

\subsection{Distance Metric Learning}

Learning an embedding space and similarity score with respect to a specific semantic label becomes useful in creating generalizable frameworks that can be used for classification as well as clustering tasks. They are more suited for open set scenarios as the scores are learned on a match vs. non-match basis and do not compute per-class probability forcing the score to be maximized for one of $n$ classes. Classification-based losses for metric learning have shown promising results in applications such as face verification, but pairwise losses are more prevalent for image retrieval scenarios~\cite{zhai2019classification, movshovitz2017no}. Intuitively, networks trained with pairwise losses are intended to generalize better. Among the large pool of such loss functions including, but not limited to, contrastive~\cite{simo2015discriminative}, triplet~\cite{kumar2016learning} and N-pair~\cite{sohn2016improved}, triplet loss is the most popular and is effective in numerous image comparison tasks. Most of these approaches learn similarity in a presumably metric space~\cite{scheirer2014good} but mostly for classification. Methods that use augmented triplet loss~\cite{xuzhang2017learningdisc, xuzhang2017learningspr} for detection and description in a transformation-aware space have inspired this work. The method proposed in this paper uses quadruplets that contain an extra weak positive sample for ordering distance between positive samples. A different quadruplet approach was employed by Chen et al.~\cite{chen2017beyond} for person re-identification and Zhang et al.~\cite{zhang2017learning} for image correspondence. Chen et al.~claim better generalization for their test sets through the use of two references, one each for the positive and negative samples in the quadruplets (set of two triplets), than triplet loss. 

\subsection{Learning to Rank}
Image ordering involves comparing all possible image pairs and then ranking them based on the distance from the reference or query. Hence, this operation is similar to 
algorithms from the information retrieval domain such as RankNet~\cite{burges2005learning} and ListNet~\cite{cao2007learning} that aim towards learning a correct ranking among a set of objects. RankNet minimizes the number of inversions in the rank as its objective whereas ListNet employs listwise loss functions. Chen et al.~\cite{chen2009ranking} explain the relationship between losses used for learning to rank and the evaluation metrics. They do so using essential loss to model ranking as a sequence of classification tasks and establish that the minimization of ranking losses lead to the maximization of evaluation measure. Our quadruplets are lists of four images where three are related through image transformations.

So far, in the image domain, learning to rank methods 
have focused more on semantically similar images versus dissimilar images rather than near-duplicates~\cite{hu2008multiple,Cakir_2019_CVPR}. Semantically similar images are images that share content but have not been derived from the same image and come from different imaging pipelines. It makes sense for applications such as content-based image retrieval to focus more on such images as they care about variety and relevance in the retrieved results. Retrieving near-duplicates would not be very useful in such scenarios as a user would want to see different images (in terms of camera, view and context) of the same content. Whereas in order to perform provenance analysis on any image, it becomes imperative to attempt to understand the near-duplicate variants at a finer granularity in terms of the image space.





\section{Transformation-Aware Distance Learning}
\label{sec:algo}
A fundamental task in provenance analysis, like
other visual recognition problems, is computing the dissimilarity between two images.
Techniques currently used in the literature, such as keypoint matching and color-wise content comparison with mutual information, present the drawback of not taking into account modifications other than affine and basic color transformations. They also do not consider complex transformations and their ordering.
Aiming to increase the reliability of dissimilarity computation between two images that share content in a way where one gives origin to the other, after a number of transformations, we propose the use of deep distance learning.

A key contribution of the proposed framework is a ranking-based network training approach that learns how to express the dissimilarity between such images as a function of the number of transformations. We describe the approach in detail in the following sections.

\subsection{Training with Quadruplets}
\label{subsec:quad}

\begin{figure}[t]
\begin{center}
\includegraphics[width=0.8\linewidth]{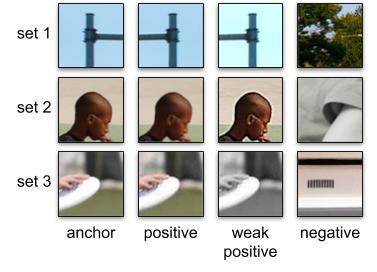}
\end{center}
\vspace{-0.5cm}
\caption{
Examples of quadruplets used for training in the proposed algorithm.
Each row depicts one set of quadruplets, in the following column-wise order: the anchor, the positive (after one transformation), the weak positive (after two transformations), and the negative (unrelated) patch.
The number of transformations is for illustration sake only; other configurations are possible (See Sec.~\ref{sec:exp}).
Anchors and negatives come from the COCO dataset~\cite{lin2014microsoft}.
}
\label{fig:patches}
\vspace{-5mm}
\end{figure}

Similar to triplet-loss training, we want to encode images in a space where, given an image of reference (\textit{i.e.}, the anchor), positively related images lie closer to it than unrelated images. For our purpose, 
we have the additional requirement of ordering the positively related images, which limits the use of conventional triplet sampling regime.
Hence, we propose learning embeddings using quadruplets to better facilitate ordering among the related images.

When training the network, we provide sets of four patches, namely (i)~the \emph{anchor} patch, which represents the original content, (ii)~the \emph{positive} patch, which stores the anchor after $M$ image processing transformations, (iii)~the \emph{weak positive} patch, which stores the positive patch after $N$ transformations, and (iv)~the \emph{negative} patch, a patch that is unrelated to the others (see Figure~\ref{fig:patches}).
The idea is to train the embedding network to provide a distance score to a given pair of patches, where the output score between the anchor and the positive patch is smaller than the one between the anchor and the weak positive, which, in turn, is smaller than the score between the anchor and the negative patch.

To obtain the quadruplets of patches for training, we employ a specific set of image transformations that are of interest to provenance analysis.
They are: (i)~projective changes (\textit{e.g.}, content scaling, rotation, flipping, shear, and projection), (ii)~color-space changes (\textit{e.g.}, changes in brightness, in contrast, gamma correction, and grayscaling), (iii)~frequency-space changes (\textit{e.g.}, blurring and sharpening), and (iv)~data-lossy compression.
For each anchor patch, random transformations from this pool are sequentially applied, one on top of the result of the other, allowing us to generate positive and weak positive patches from the anchor, after $M$ and $M+N$ transformations, respectively.
Figure~\ref{fig:patches} depicts a few examples of these quadruplets.

\begin{figure*}[t]
\begin{center}
\includegraphics[width=0.9\linewidth]{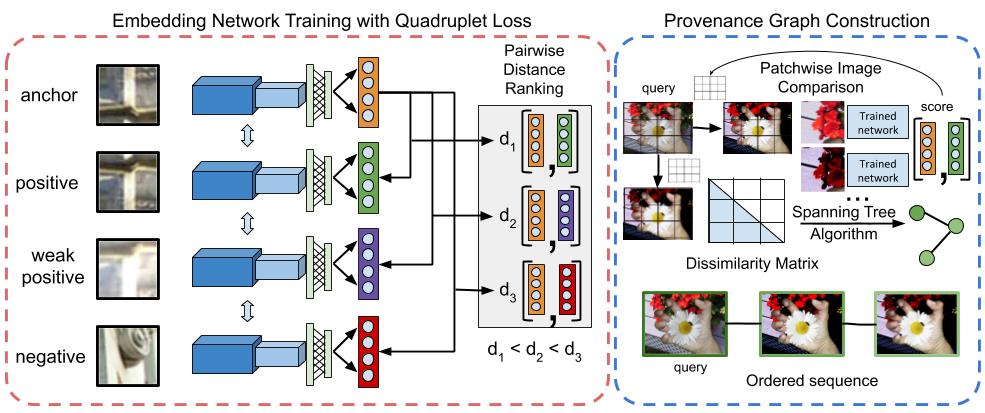}
\end{center}
  \caption{The left box (red) shows the proposed framework used to train a network that can provide transformation aware embeddings to input regions.  
  The right box (blue) shows steps for how the trained embedding network is used to construct provenance graphs. Once the feature extractor is trained, we describe image regions from the test set, use L2-distance to compute pairwise image dissimilarity matrices and use a greedy approach to connect images into a provenance graph.} 
\label{fig:pipeline}
\vspace{-3mm}
\end{figure*}

\begin{table}[]
    \centering
    \caption{Details of the architecture of the embedding network. The convolutional layer structure is defined as $C \times H \times W(Padding)$. The dimensionality of the learned feature for each patch is 256. The number of trainable parameters in the network is 7,564,800.}
    \vspace{0.3cm}
    \begin{tabular}{|l|c|c|c|}
    \hline
    Layer & Structure & Output & Parameters \\\hline\hline
        Conv1+bn & 32x11x11 (9) & 72x72 & 11,712\\\hline
        MaxPool & (2,2) & 36x36 & --\\\hline
        Conv2+bn & 64x9x9 (7) & 42x42 & 166,080 \\\hline
        MaxPool & (2,2) & 21x21 & --\\\hline
        Conv3+bn & 128x7x7 (5) & 25x25 & 401,792 \\\hline
        MaxPool & (2,2) &  & --\\\hline
        Conv4+bn & 256x5x5 (3) & 14x14 & 819,968 \\\hline
        MaxPool & (2,2) &  & --\\\hline
        Conv5+bn & 512x3x3 (1) & 7x7 & 1,181,184 \\\hline
        MaxPool & (2,2) &  & --\\\hline
        Linear & -- & 1024 & 4,719,616\\\hline
        BatchNorm1d & -- & 1024 & 2,048 \\\hline
        Linear & -- & 256 & 262,400\\\hline
    \end{tabular}
    \vspace{-3mm}
    
    \label{tab:network}
\end{table}

\subsection{Network, Loss Function and Optimization}
The network used to learn representations is a four-way Siamese structure with all four branches of the network sharing weights (see the training block of Figure~\ref{fig:pipeline}).
Each embedding module is a five-layer convolutional neural network (CNN) with one batch normalization layer and two fully connected layers, very similar to the one used in~\cite{chen2017image} to identify image processing operations. In contrast to them, we do not preprocess the images using high-pass filters and let the network learn useful noise patterns depending on the transformation. Our convolutional layers use Rectified Linear Unit (ReLU) as activation function for faster convergence. Max-pooling is applied to feature maps from convolutional layers. We use relatively large kernels in our convolutional layers to have a larger receptive field for our learned feature maps. Details of the network architecture are provided in Table~\ref{tab:network}.  The features extracted from the embedding units are used by the pairwise distance ranking unit to compute the loss. 

We employ an L2-distance-based pairwise margin ranking loss to learn image embeddings that match the order of generation of near-duplicate images through edits and transformations. Given a feature vector for 
an anchor image
patch $a$ and two transformed derivatives of the anchor patch $p$ (positive) and $p^\prime$ (weak positive) where $p=T_M(a)$ and $p^\prime=T_N(T_M(a))$, and an unrelated image patch from a different image $n$, the objective is to learn embeddings such that $d(a,p) < d(a,p^\prime) < d(a,n)$, a quadruplet based similarity precision rank~\cite{wang2014learning}. Here, $d(.)$ is the L2-distance between two embeddings. $T_M$ is the series of $M$ transformations applied to $a$ to generate $p$ and $T_N$ is the series of $N$ transformations further applied on $p$ to create $p^\prime$. 

In order to achieve the rank objective, we minimize:


\vspace{-0.5cm}
\begin{multline}
    {\mathcal L}(a, p, p^\prime, n) = \\
    \mathrm{max}(0, -y\times(d(a, p^\prime) - d(a, n)) + \mu_1) + \\
    \mathrm{max}(0, -y\times(d(p, p^\prime) - d(p, n)) + \mu_2) + \\
    \mathrm{max}(0, -y\times(d(a, p) - d(a, p^\prime)) + \mu_3)\ \ \ \ \ \ \ \ 
    \label{eqn:mrlquadruplet}
\end{multline}
In the above loss function, $y$ is the truth function (analogous to labels for classification) which determines the rank order~\cite{rudin2009margin} and $\mu_1$, $\mu_2$ and $\mu_3$ are margins corresponding to each pairwise distance term and are treated as hyperparameters. The ranking operates in the score space rather than the embedding space. Also, $\mu_1 > \mu_2 > \mu_3$, in order to cluster stronger positives (less transformed versions) closer to the anchor image embedding. Values for these margins are chosen empirically over a small range of proportional values $\in$ [0.01, 0.1]. Here, proportion is based on the fact that distances are L2 normalized and $d(a,n) \gg \{d(a,p), d(a,p^\prime)\}$ is desirable. Margin ranking loss determines the loss value based on 
the sign of $y$ and the difference between the two distances. 
Both having the same sign implies ordering is correct and the loss is zero. A positive loss is accumulated when the ordering is wrong and they are of opposite signs. 
The combination of these terms induce an ordering
on the four units in the quadruplet.

We use three out of the four possible topologically ordered triplets from the quadruplets, \textit{i.e.}, $(a, p, p^\prime)$, $(a, p^\prime, n)$ and $(p, p^\prime, n)$ and claim that they are enough to rank the involved pairwise distances in an increasing order. The fourth triplet $(a, p, n)$ is redundant as the ordering of pairs $(a, p)$ and $(a, n)$ is already covered by second and third terms of the loss by transitivity, \textit{i.e.}, if $d(a, p) < d(a, p^\prime)$ and $d(a, p^\prime) < d(a, n)$, then $d(a, p) < d(a, n)$. Minimizing the proposed loss maximizes the similarity precision as the first distance in each of the three terms of the loss is maximally bounded by the second distance in that term, leading to the desired quadruplet ordering $d_1 < d_2 < d_3$ (see Figure~\ref{fig:pipeline}).



The above loss is optimized using Stochastic Gradient Descent method with Nesterov Momentum~\cite{sutskever2013importance}. The learning rate is reduced by a factor of $0.1$ every time the validation loss plateaus. The network is trained until convergence (usually around 100 epochs). The model is saved for epochs where the average similarity precision for the validation set of quadruplets improves over the previous best model. The model corresponding to the best measure for validation is used for feature extraction from patches of test images. 

\subsection{Image Provenance Graph Construction}
\label{subsec:prov}

Features learned with the proposed technique are used to perform image comparison for image provenance analysis. 
For this paper, we assume that the related set of images for a given image is provided to us through an accurate retrieval method similar to the \textit{oracle} scenario described in~\cite{bharati2017uphy, moreira2018image}. 
Once a set of $k$ images related to a query image is obtained, provenance graph construction involves the following 
steps:

\textit{1. Creation of dissimilarity matrix}. All possible image pairs from the retrieved image set are compared to obtain a similarity or distance score that is used to create a dissimilarity matrix $\mathcal{D}$ of size $k \times  k$. Depending on the properties of the dissimilarity measures, the matrix can be symmetric or asymmetric. In order to use the 
proposed framework for obtaining pairwise image dissimilarity scores, we extract patches of the same size as those in the quadruplets used for training from each of the $k$ images. These patches are then described using the trained embedding module of the proposed framework. $\mathcal{D}[i,j]$ between image $i$ and image $j$ is computed by matching the set of features from one image to the other using an all-to-all brute force matching strategy. We only consider the bidirectionally consistent top match for each patch and compute the average L2-distance for all the patches with their best matches. This is computed for each of the $\binom{k}{2}$ possible image pairs to get a symmetric dissimilarity matrix $\mathcal{D}$.

\textit{2. Combining image pairs to construct a graph}. Considering the matrix $\mathcal{D}$ as an adjacency matrix with the dissimilarity values as edge weights of a fully connected graph, Kruskal's optimum spanning tree algorithm~\cite{Kruskal:AMS:1956} is employed to connect images in a greedy manner. Since we use L2-distance between embeddings as dissimilarity values, the algorithm chooses edges with lower weights first and adds $(n-1)$ such edges until all the $n$ images are connected. The result is an undirected ordered graph with $n$ related images as nodes and each selected edge corresponds to the image transformation between the two connected vertices.
    
We focus on the graph construction part of image provenance analysis as it entails reconstructing a contiguous sequence of image transformations (similar to the output chain in Figure~\ref{fig:teaser}). 
As the proposed approach is agnostic to the directionality of transformation between a given pair of images, for the purposes of this work, we create and evaluate the undirected provenance graphs similar to Bharati et al.~\cite{bharati2017uphy}. An asymmetric distance function that correctly infers whether an operation was performed from image 1 to image 2 or a reverse operation from image 2 to 1 based solely on visual content is hard to compute, as pointed out in the literature~\cite{bharati2019beyond}. 
Since we employ a greedy approach to connect images for provenance graph building, the ordering between pairwise distances is still important as it governs the choice of image pairs that will share a direct vs. an indirect relationship in the graph. 

\section{Experiments}
\label{sec:exp}

The transformation-aware embedding model is trained using $\sim$350k patches extracted from 4,416 images selected from the COCO 2017 unlabeled partition~\cite{lin2014microsoft}.
Each image is synthetically modified with a sequence of transformations randomly selected from the pool described in Sec.~\ref{subsec:quad}, with no repetitions.
We prepare easy and hard sets of quadruplets, which are equally sampled to compose the training batches.
Easy quadruplets have a larger number of transformations between positive and weak positive patches, namely one and four transformations on top of the anchor, or two and five transformations.
Hard quadruples, in turn, have a smaller number of transformations in between, namely one and two, or two and three transformations on top of the anchor.
Patches with size $64 \times 64$ pixels are sampled centered on SURF keypoints, that are previously described and tracked across the original and transformed images.
Negative patches have the same size and are obtained from keypoints detected over unrelated images.


During evaluation, $64 \times 64$ patches are sampled from the images and described with the trained network, leading to 256-dimensional feature vectors whose pairwise L2-distances are used to create a dissimilarity matrix as explained in Sec.~\ref{subsec:prov}.
Once the dissimilarity matrix is computed, 
undirected graphs are generated using Kruskal's minimum (or maximum depending on if comparison score is a distance or similarity) spanning tree algorithm.
Code for the full pipeline to perform provenance graph construction using transformation-aware learned embeddings 
will be released to the community, upon the acceptance of this paper for publication.


\subsection{Datasets}
The DARPA Media Forensics program has curated and released multiple datasets for image provenance analysis as part of its annual challenges. These datasets  contain good quality images of generic scenes, along with corresponding manipulated versions with a record of each step of manipulation and intermediate versions in the form of graph journals. The manipulations were performed using GIMP, an open source image editing tool, by professional  artists~\cite{guan2019mfc}. The journals capture a vast array of image manipulation operations from color-based transformations to anti-forensic operations. We evaluate our method on two DARPA datasets:

   \textit{1. Nimble 2017 Challenge Dataset.} We use the NC2017-Dev1-Beta4 set~\cite{nist2017dataset}, which contains provenance journals for 65 queries. Similar to the experimental setup described in~\cite{moreira2018image}, we also the \emph{development} partition of this dataset since it provides a full set of ground-truth graphs. The order of the ground-truth graphs is in the range $[2,81]$, with the average graph order being 13.6. The resolution of images in the dataset varies widely, with the average resolution being 5.9 megapixels.
   
   \textit{2. Media Forensics 2018 Challenge Dataset.} We use the MFC18-Dev1-Ver1~\cite{guan2019mfc} partition of the 2018 challenge dataset. This partition has provenance journals related to 258 query images. The average graph order is 14.3 and the average resolution of images is 10.1 megapixels. The distribution of manipulations in this dataset is different from those in  NC2017-Dev1-Beta4~\cite{guan2019mfc}. These provenance cases have a larger set of manipulations. For example, operations such as recapture and \textit{CGI-Fill} are not present in the NC2017-Dev1-Beta4 dataset. Please check the supplemental material for more details on the available manipulations in the dataset.

The DARPA Media Forensics challenges and evaluation protocols for image provenance analysis contain two tasks: (1) an end-to-end analysis, and (2) an oracle framework. The first protocol includes retrieval of related images given a query and construction of final provenance graphs. This task includes distractors in the pool of world images along with those related to the query. The oracle framework provides a standalone estimate of the effectiveness of the graph construction solution where errors from the previous step are not taken into account. For this work, we assume that we have access to all of the related image components for performing provenance analysis for a query image because our focus is ordering of near-duplicate images.

\subsection{Baseline Approaches}
 The first category of comparison methods incorporates detection of salient keypoints in the images and matches local regions around those points with those in other images. For experiments on NC17-Dev1-Beta4, we consider SURF~\cite{bay2006surf}, LIFT~\cite{yi2016lift}, DELF~\cite{noh2017large} and DeepMatching~\cite{revaud2016deepmatching}, a hierarchical dense local region matching technique. SURF and DeepMatching are handcrafted while LIFT and DELF are learned local features. For the LIFT experiment, we use the published model trained on the Piccadilly Circus images~\cite{wilson_eccv2014_1dsfm}. The model used for DELF was trained on the Google Landmarks dataset~\cite{GLandMarks}. After matching, we perform geometric consistency verification~\cite{bharati2017uphy} for matches from 2000 keypoints for SURF, 500 for LIFT, and all detected regions for DELF. DeepMatching inherently performs this verification step by design, enabling us to use the matches directly.
The number of consistent matches between two images is used as 
score in the dissimilarity matrix, which is then used to create the graphs.

The second set of methods are popular data-driven learned descriptors using deep convolutional neural networks such as AlexNet~\cite{krizhevsky2012imagenet} and ResNet (18-layers)~\cite{he2016deep}, both trained on ImageNet dataset~\cite{imagenet_cvpr09}. For these, we describe $64\times64$ patches extracted from the entire image. We then match pairs of patch sets between images as described in Section~\ref{subsec:prov}, and create the dissimilarity matrix. The graphs are created in the same manner for all the methods in consideration.
We later choose the best performing methods from the different types to evaluate performance on MFC18-Dev1-Ver1 and report results using the same steps as used for NC17-Dev1-Beta4.

\subsection{Evaluation}
During training, the 4-way network (see training panel in Figure~\ref{fig:pipeline}) is evaluated using similarity precision for quadruplets, similar to the usage for triplets in~\cite{wang2014learning}. This metric is defined as the percentage of quadruplets in the validation set for which the pairwise distances are correctly ordered as $d(a, p) < d(a, p') < d(a, n)$. 
To evaluate the proposed framework for provenance graph construction, we generate graphs by creating dissimilarity matrices using distance scores between learned embeddings and Kruskal's optimum spanning tree algorithm. We compare them to the ground-truth graphs for the datasets
using their metrics for the provenance task~\cite{nist2017plan}. 

The metrics are computed by comparing the nodes and edges from both ground-truth and candidate graphs. The corresponding measures of  \textbf{Vertex Overlap (VO)} and \textbf{Edge Overlap (EO)} are the harmonic mean of precision and recall (F1 score) for the graph vertices and edges retrieved by our method. In addition to these, a unified metric representing one score for the graph overlap namely the \textbf{Vertex Edge Overlap (VEO)} is also reported. The VEO is the combined F1 score for vertices and edges. All the metrics are computed through the \emph{MediScore} tool~\cite{mediscore}, with \textit{undirected graph} option.
The values of these metrics lie in the range $[0,1]$ where higher values are better. 






\section{Results}

\begin{table*}[t]
\renewcommand{\arraystretch}{1.4}
\footnotesize
\centering
\caption{
Provenance graph construction over the NC2017-Dev1-Beta4 dataset (oracle mode).
We report the mean and the standard deviation of 65 cases for the metrics presented in Sec.~\ref{sec:exp}.
In bold: best results. TAE stands for Transformation Aware Embeddings learned using the proposed approach. NA stands for \textit{not applicable}. Please see supplemental material for qualitative results.
}
\vspace{0.2cm}
\begin{tabular}{L{2.2cm}L{1.4cm}L{1.7cm}R{1.7cm}R{1.6cm}R{1.8cm}R{1.8cm}R{1.8cm}}
\hline
Description Method & Description Type & Local Feature Type & Feature Vector~Size (\#) & Disk Size (MB) & \multicolumn{1}{c}{VO} & \multicolumn{1}{c}{EO} & \multicolumn{1}{c}{VEO} \\
\hline
SURF~\cite{bay2006surf} & handcrafted & keypoints & 64 & 851 & 0.90 ($\pm$0.08) & 0.65 ($\pm$0.16) & 0.78 ($\pm$0.11) \\
LIFT~\cite{yi2016lift} & learned & keypoints & 128 & 89 & 0.79 ($\pm$0.18) & 0.39 ($\pm$0.23) & 0.60 ($\pm$0.19) \\
DELF~\cite{noh2017large} & learned & keypoints & 40 & 205 & 0.86 ($\pm$0.18) & 0.59 ($\pm$0.23) & 0.73 ($\pm$0.19) \\
DeepMatching~\cite{revaud2016deepmatching} & handcrafted & keypoints & NA & 230 & 0.59 ($\pm$0.37) & 0.28 ($\pm$0.25) & 0.44 ($\pm$0.30) \\
AlexNet~\cite{krizhevsky2012imagenet} & learned & image patches & 4096 & 19000 & 1.00 ($\pm$0.00) & 0.61 ($\pm$0.15) & 0.81 ($\pm$0.08) \\
ResNet-18~\cite{he2016deep} & learned & image patches & 512 & 2400 & 1.00 ($\pm$0.00) & 0.61 ($\pm$0.17) & 0.81 ($\pm$0.08) \\
\textbf{TAE (ours)} & \textbf{learned} & \textbf{image patches} & \textbf{256} & \textbf{1200} & \textbf{1.00 ($\pm$0.00)} & \textbf{0.66 ($\pm$0.14)} & \textbf{0.83 ($\pm$0.07)} \\
\hline
\end{tabular}
\label{tab:results_nc17}
\end{table*}

\begin{table}[t]
\renewcommand{\arraystretch}{1.4}
\footnotesize
\centering
\caption{
Provenance graph construction over the MFC18-Dev1-Ver1 dataset (oracle mode).
Means and standard deviations of 258 cases are reported for the metrics presented in Sec.~\ref{sec:exp}.
Only the best deep learning-based keypoint and full-image approaches from Table~\ref{tab:results_nc17} are reported here. 
In bold: best results. 
}
\vspace{0.2cm}
\begin{tabular}{L{1.8cm}R{1.5cm}R{1.5cm}R{1.5cm}}
\hline
Description & \multicolumn{1}{c}{VO} & \multicolumn{1}{c}{EO} & \multicolumn{1}{c}{VEO} \\
\hline
SURF~\cite{bay2006surf} & 0.89 ($\pm$0.17) & 0.65 ($\pm$0.20) & 0.78 ($\pm$0.17) \\
DELF~\cite{noh2017large} & 0.84 ($\pm$0.22) & 0.59 ($\pm$0.25) & 0.72 ($\pm$0.22) \\
ResNet-18~\cite{he2016deep} & 0.99 ($\pm$0.01) & 0.62 ($\pm$0.17) & 0.82 ($\pm$0.08) \\
\textbf{TAE (ours)} & \textbf{0.99 ($\pm$0.01)} & \textbf{0.67 ($\pm$0.17)} & \textbf{0.84 ($\pm$0.08)} \\
\hline
\end{tabular}
\label{tab:results_mfc18}
\vspace{-5mm}
\end{table}

Experiments for provenance graph construction show the efficacy of the proposed approach in comparison to existing alternatives. For each line of comparison presented, only the image correspondence measure is changed while the remainder of the provenance analysis pipeline is kept the same. 
Results presented in Tables~\ref{tab:results_nc17}~and~\ref{tab:results_mfc18} show that employing transformation-aware descriptors for image dissimilarity computation improves the average Vertex Overlap for provenance analysis. The CNN-based descriptors adapted from the area of object recognition have an advantage over keypoint-based approaches as the images that do not share a sufficient number of keypoint matches with others may not be included in the final connected graph. This leads to better node overlap for learned techniques where an all-patch-to-all-patch matching strategy is employed. That said, one of the reasons behind the popularity of keypoint-based approaches for provenance-type applications is their efficiency and subsequent ability to scale to larger problems. In this regard, the handcrafted SURF approach has an edge over the learning-based keypoint approaches (DELF and LIFT). Describing 2000 SURF keypoints for an image on average takes less than 1 second, whereas it takes $\sim$1 minute and $\sim$3 minutes for DELF and LIFT 
respectively (tested on a set of 100 images with average size of 1MB).


Improving edge overlap for provenance graph algorithms is trickier as the wrong selection of one direct edge adds a heavy penalty on the metric. It implies that the solution has a false positive edge and multiple false negative errors (equal to the number of intermediate edges that connect the related images in the groud-truth).
The observed measures of edge overlap in our results reveal that using transformation-aware image descriptors improves ordering and selection of edges over general deep learning-based descriptors. Despite being trained with less image data by an order of magnitude, the network used in this paper is more effective for the provenance graph building task. In addition to being more effective for provenance-based vertex and edge selection, our method is more efficient in terms of number of training parameters and disk space. In comparison to approximately 61M parameters for AlexNet~\cite{krizhevsky2012imagenet} and 
11M parameters for ResNet-18~\cite{he2016deep}, the embedding network used in the proposed approach contains only $\sim$ 7.5M parameters.

Another important improvement achieved by the pipeline proposed in this paper is the lowest standard deviation in terms of all the three metrics of evaluation. Since most provenance cases are quite unique with respect to image content and applied transformations, the variability in the ground-truth of provenance analysis examples is very large. Even though some images can share partial provenance lineage, it is close to impossible for two different manipulated images on the web to have undergone the same set of intermediate manipulation steps. This is why having a low standard deviation implies good generalizability of the proposed approach, which is very desirable in applications such as provenance analysis. While discussing generalizability, it is also important to note that all the models used for comparison including ours have been trained on datasets different from the test set. Thus, these experimental results also highlight the generalization capabilities of these approaches in terms of operational scenarios different from training, but still at a task that 
leverages their image understanding capabilities. 

\section{Discussion}

Estimating the order of manipulations performed on an image to generate the final version is known to be a difficult and open problem.  This paper provides a solution for ordering in the context of image provenance analysis for forensics with knowledge of a limited set of transformations during training --- the most realistic scenario one can consider.  All of our results have been evaluated for cross-dataset scenarios, and the overall design is meant for real-world operation. The evaluation set we made use of has a large set of transformations, most of which are unseen at training time. Thus we were able to reveal the limits and potentials of our approach and other possible solutions in a practical setting. 


Our efforts also highlight the need for further improvements in vision-based provenance analysis. A better understanding of the image transformation space and datasets with varied types of complex manipulations will help in the development of more advanced algorithms. In this regard, it is important to note that due to the common usage of proprietary image editing tools and the possibility of large numbers of edits, creating ground-truth for provenance analysis is an arduous undertaking.  However, through the recent efforts of the DARPA Media Forensic program~\cite{darpa}, a viable regime for data collection, annotation, and evaluation has emerged. Finally, considering the nature of the problem, it is insufficient to train a good system solely based on the data from individual images in isolation. 
The transfer of knowledge gained from the general space of image transformations is still essential to solving the problem, and the work proposed in this paper aims to advance research activity in this direction.

\section{Acknowledgement}
This material is based on research sponsored by DARPA and Air Force Research Laboratory (AFRL) under agreement number FA8750-16-2-0173. Hardware support was generously provided by the NVIDIA Corporation. We also thank the financial support of FAPESP (Grant 2017/12646-3, D\'ej\`aVu Project), CAPES (DeepEyes Grant) and CNPq (Grant 304472/2015-8).


{\small
\bibliographystyle{ieee_fullname}
\bibliography{egbib}
}

\end{document}


\title{Supplemental Material for Learning Transformation-Aware Embeddings for Image Forensics}

\maketitle
\vspace{-1.0cm}
\section{Distribution of Image Manipulation Operations in the Provenance Analysis Datasets}
\vspace{-1.cm}
\begin{figure*}
\begin{center}
\includegraphics[width=0.95\linewidth]{latex/figures/nc17_numb.pdf}
\end{center}
\vspace{-1.2cm}
\begin{center}
\includegraphics[width=0.95\linewidth]{latex/figures/mfc18.pdf}
\end{center}
\vspace{-0.5cm}
\label{fig:dataset}
\end{figure*}

\newpage
\section{Possible Triplets from a Quadruplet}

\begin{figure*}[htb]
\vspace{-0.5cm}
\begin{center}
\includegraphics[width=0.5\linewidth]{latex/figures/PossibleTriplets.pdf}
\end{center}
 \vspace{-0.5cm}
  \caption{Given an ordered quadruplet $(a, p, p^\prime, n)$, there are four possible ordered triplets $(a, p, p^\prime)$, $(a, p, n)$, $(p, p^\prime, n)$ and $(a, p^\prime, n)$ as shown above. Any other possible permutation of a triplet will violate the expected linear ordering of the distance functions.}
 
\label{fig:triplets}
\end{figure*}

\vspace{-0.5cm}
\section{Qualitative Results for Provenance Graph Construction}
\subsection{Case 1}

\begin{figure}[!htb]
\begin{center}
\begin{subfigure}{0.8\textwidth}
\centering
\includegraphics[width=\textwidth]{latex/figures/gt1.pdf}
  \caption{Ground truth provenance graph (contd.)}
\label{fig:gt1}
\end{subfigure}
\end{center}
\end{figure}
\begin{figure}[!htb]\ContinuedFloat
\begin{center}
\begin{subfigure}{0.8\textwidth}
\centering
\includegraphics[width=\textwidth]{latex/figures/tadl1.pdf}
  \caption{Graph generated using the proposed Transformation Aware Embeddings (contd.) }
\label{fig:tadl1}
\end{subfigure}
\end{center}
\end{figure}
\begin{figure}[!htb]\ContinuedFloat
\begin{center}
\vspace{-0.2cm}
\begin{subfigure}{0.8\textwidth}
\centering
\includegraphics[width=\textwidth]{latex/figures/surf1.pdf}
  \caption{Graph generated using SURF descriptions}
\label{fig:surf1}
\end{subfigure}
\vspace{0.2cm}
\begin{subfigure}{0.8\textwidth}
\centering
\includegraphics[width=\textwidth]{latex/figures/resnet1.pdf}
  \caption{Graph generated using patch-based descriptions from ResNet18}
\label{fig:resnet1}
\end{subfigure}
\vspace{-0.2cm}
\caption{Green edges are correct while red edges are false positives. VertexEdgeOverlap for the proposed solution, SURF and ResNet18 are 0.83, 0.72 and 0.8 while the EdgeOverlap is 0.66, 0.61 and 0.59 respectively. Due to fewer keypoints shared by donor images, the SURF solution misses 4 nodes or images in graph construction.}
\end{center}
\end{figure}

\afterpage{\clearpage
\subsection{Case 2}

\begin{figure}[!htb]

\begin{subfigure}{0.3\textwidth}
\centering
\includegraphics[height=10cm]{latex/figures/gt2.pdf}
  \caption{Ground truth provenance graph}
\label{fig:gt2}
\end{subfigure}
\begin{subfigure}{0.35\textwidth}
\centering
\includegraphics[height=10cm]{latex/figures/tadl2.pdf}
  \caption{Graph generated using the proposed Transformation Aware Embeddings}
\label{fig:tadl2}
\end{subfigure}
\begin{subfigure}{0.35\textwidth}
\centering
\includegraphics[height=10cm]{latex/figures/surf2.pdf}
  \caption{Graph generated using SURF descriptions}
\label{fig:surf2}
\end{subfigure}
\caption{Green edges are correct while red edges are false positives. VertexEdgeOverlap for the proposed solution and SURF are 0.91 and 0.9 while the EdgeOverlap is 0.8 and 0.88 respectively. Due to low robustness of matched keypoints, the SURF solution misses 1 node (image) in the graph.}

\end{figure}
}